%% file: main.tex
\documentclass[conference]{IEEEtran}
\IEEEoverridecommandlockouts
\usepackage{lipsum}
\usepackage{mwe}
\usepackage{mathtools}
\usepackage{makecell}
\usepackage{cite}
\usepackage{amsmath,amssymb,amsfonts}
\usepackage{algorithmic}
\usepackage{graphicx}
\usepackage{textcomp}
\usepackage{xcolor}
\usepackage{wrapfig}
\usepackage{multirow}
\usepackage{tabularx,booktabs}
\usepackage[super]{nth}
\usepackage{tikz}
\def\BibTeX{{\rm B\kern-.05em{\sc i\kern-.025em b}\kern-.08em
    T\kern-.1667em\lower.7ex\hbox{E}\kern-.125emX}}
    
\begin{document}

\title{Facetron: A Multi-speaker Face-to-Speech Model \\ based on Cross-Modal Latent Representations}
\author{Seyun Um, Jihyun Kim, Jihyun Lee, Hong-Goo Kang\\
\IEEEauthorblockA{\textit{Dept. of Electrical and Electronic Engineering} 
\textit{Yonsei University}
Seoul, Korea }
}

\maketitle

\begin{abstract}
In this paper, we propose a multi-speaker face-to-speech waveform generation model that also works for unseen speaker conditions.
Using a generative adversarial network (GAN) with linguistic and speaker characteristic features as auxiliary conditions, our method directly converts face images into speech waveforms under an end-to-end training framework.
The linguistic features are extracted from lip movements using a lip-reading model, and the speaker characteristic features are predicted from face images using a face encoder trained through cross-modal learning with a pre-trained acoustic model.
Since these two features are uncorrelated and controlled independently, we can flexibly synthesize speech waveforms whose speaker characteristics vary depending on the input face images.
We show the superiority of our proposed model over conventional methods in terms of objective and subjective evaluation results. 
\footnote{The demo samples of the proposed and other models are available at https://sam-0927.github.io/}
\end{abstract}


\input{1_introduction}

\input{2_related_work}

\input{3_proposed_model}
\input{4_training_criterion}

\input{5_experiment}

\input{6_ablation}
\input{7_conclusion}

\bibliographystyle{IEEEbib}
\bibliography{reference}

\end{document}

%% file: 1_introduction.tex
\section{Introduction}
\label{sec:intro}
Automated lip-reading (ALR) is the task of predicting linguistic information from a sequence of visual lip movements. 
It has become an essential technology for communications for hearing-impaired people, silent communications for keeping privacy, and so on~\cite{assael2016lipnet, chung2017lip}. 
Especially, there have been several attempts to generate high-quality speech waveforms directly from lip movements and face images based on deep learning frameworks~\cite{vougioukas2019videodriven, prajwal2020learning, goto2020face2speech, ephrat2017vid2speech}. 

To synthesize human voices from silent lip movements and face images, one simple approach is to concatenate a text-to-speech (TTS) module to a module that converts either face or lip images to contextual variables.
However, this approach has a limitation in that unavoidable recognition errors at the contextual conversion module degrade the synthesis performance of the following speech synthesis module. 
Thus, there have been several attempts to estimate acoustic information rather than phonetic information as intermediate features for waveform synthesis, which have been shown to improve the robustness of the whole synthesis framework. 
These acoustic features include vocoding parameters such as pitch, voicing information, and spectral parameters. 

Meanwhile, Vougioukas \textit{et al.}~\cite{vougioukas2019videodriven} proposed a direct generation method of speech waveforms from a sequence of lip images based on GAN framework. 
Though they were able to generate more realistic and intelligible speech in both speaker-dependent and speaker-independent settings thanks to the GAN approach, there is still room for improvements of the quality of the generated speech.

In this paper, we use a GAN-based method to generate raw speech waveforms from lip and face image sequences. 
We obtain two intermediate features: contextual information from lip movements and speaker characteristic-related features from face images. 
To estimate the speaker characteristic-related acoustic features from face images, we introduce a cross-modal learning method that utilizes the cosine similarity between embeddings obtained from face images and prosody embeddings obtained from a pre-trained prosody-extracting module~\cite{valle2020mellotron}. 
Both contextual and speaker-related acoustic embeddings are used as conditional features for a GAN-based waveform generation module.
Based on various experiments using small and large-scale datasets, we show that our model successfully generates high-quality speech even for unseen speakers.

Our main contributions of this paper are as follows:
1) We introduce various training criteria to improve the quality of synthesized speech in a multi-speaker face-to-speech generation scenario.
2) Our model extracts linguistic and speaker characteristic-related embeddings independently; thus, it is possible to modify speaker characteristic information without changing the linguistic information.
3) To reliably extract speaker characteristic-related information, we propose an effective cross-modal learning technique using a pre-trained prosody model.
Since the prosody model was pre-trained on a database with a large number of speakers, we can successfully predict the acoustic information for many speakers even if they were unseen during training. 
4) We use a well-established GAN-based generator to synthesize high-fidelity speech from facial movements. Notably, the discriminator module includes an effective discrimination criterion motivated by a perceptual quality metric, which induces the generated speech to become more natural.

%% file: 2_related_work.tex
\section{Related Work}
\label{sec:related_work}
\subsection{Lip-to-speech synthesis}

Lip-to-speech algorithms consist of two main stages: the extraction of intermediate features from input images and the generation of speech waveforms from them.
Vid2Speech~\cite{ephrat2017vid2speech} and Lipper~\cite{kumar2019lipper} extract line spectrum pairs (LSPs) from face images and synthesize speech waveforms using linear prediction (LP) synthesis. 
However, since low order LP coefficients do not include sufficient acoustic information for speech synthesis, the generated speech sounds are often not natural.
Vid2voc~\cite{michelsanti2020vocoder} adopts the WORLD vocoder~\cite{morise2016world} to improve speech quality. To further improve intelligibility, this method jointly trains a visual speech recognition network and the vocoder parameter prediction network using a multi-task learning framework.

Another line of approaches is predicting spectrograms or mel-spectrograms, which contain more detailed acoustic information, and using them as intermediate features for speech waveform generation.
One example of this approach is an extension of Vid2Speech~\cite{ephrat2017vid2speech} which uses facial movements as additional inputs to predict such spectrograms.
Another example, Lip2AudSpec~\cite{akbari2018lip2audspec}, uses a pre-trained autoencoder to compress input embeddings and reconstruct spectrograms. By utilizing a well-established sequence-to-sequence architecture,
Prajwal \textit{et al.}~\cite{prajwal2020learning} develop Lip2Wav that learns the speaking styles of a small number of target speakers.
Vougioukas \textit{et al.}~\cite{vougioukas2019videodriven} directly generate raw speech waveforms using adversarial training, but their improvements are not particularly significant because they use a simple architecture of a stack of transposed convolutions for a generator module.

Our proposed model defines linguistic and acoustic features as intermediate features.
We extract those features using two feature extractors: a lip-reading network that predicts linguistic features and a face encoder that predicts speaker characteristics. 
We also utilize the high-synthesis ability of a GAN-based neural vocoder to generate natural speech waveforms.

\subsection{Face-speech cross-modal learning}

Cross-modal features between face and speech have been used to improve the performance of various speech-related tasks such as speech separation, speech enhancement, and speaker recognition. 
Due to their matching correspondence, the features can be also used for a domain transformation task between face and speech.
For example, Speech2Face~\cite{oh2019speech2face} synthesizes a face image given speech segments as input. 
In Face2Speech~\cite{goto2020face2speech}, a pre-trained multi-speaker TTS system synthesizes speech given speaker embeddings extracted from face images.
Our method uses a metric-learning approach to train our face encoder to predict similar speaker characteristics from the features extracted from face images.

%% file: 3_proposed_model.tex
\section{Proposed Model}
\label{sec:proposed_model}

\begin{figure}[!t]
\centering
\includegraphics[width=0.8\linewidth]{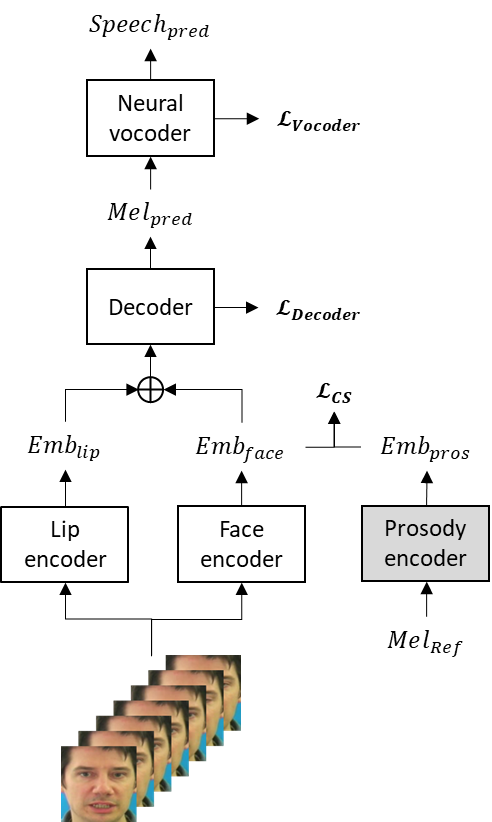}
\caption{\footnotesize System architecture of the proposed model. The input to the waveform generator (HiFi-GAN) is a concatenation of face embedding and lip embedding generated by the face encoder and the lip encoder, respectively. Gray block refers to pre-trained module.}
\label{fig:architecture}
\end{figure}

Figure \ref{fig:architecture} illustrates the system architecture of our proposed model, which consists of a mel-spectrogram generator and a neural vocoder. Our mel-spectrogram generator, which we call Facetron, transforms face and lip images into mel-spectrograms. It consists of three modules: a lip encoder, a face encoder, and a decoder. We then use a high-quality neural vocoder HiFi-GAN~\cite{kong2020hifi} to synthesize waveforms from the generated mel-spectrogram.
Unlike a TTS model which has a one-to-many mapping problem, Facetron doesn't suffer from the mismatch of time alignment because the frame rate of face images is fixed.
Detailed descriptions of each module in Facetron are provided in the following paragraphs.

\noindent\textbf{Lip encoder.}
Lip encoder is a feature extractor that outputs lip embeddings given a sequence of lip images with size $144 \times 144$ cropped from face images.
The lip encoder is able to predict contextual and linguistic information because it is trained with additional full-connected layers which predict actual graphemes from the output of the lip encoder.
The design of the lip encoder is based on LipNet~\cite{assael2016lipnet}.
It consists of a 3-layer 3D CNNs, a residual block with channel-wise dropout and spatial pooling, a 2-layer bidirectional gated recurrent unit (Bi-GRU), and 2 fully connected layers.
The residual block consists of two 3D CNNs that have 96-32 channels with the same kernel size of (3, 3, 3), a stride size of (1, 1, 1), and a padding size of (1, 1, 1). These bottleneck blocks play an important role in improving performance by extracting linguistic features.
The output is multiplied by a scale factor of 0.2. Then, all outputs of 3D CNNs blocks are fed to the Bi-GRU module.
We use connectionist temporal classification (CTC) loss~\cite{graves2006connectionist} as the criterion for recognizing characters.

\vspace{2pt}
\noindent\textbf{Face encoder.}
Our proposed model targets a multi-speaker system, which can synthesize different speech depending on speaker identities even when the linguistic information is the same.
Thus, we define face embedding, which is the output of face encoder, to represent speaker characteristics and use it as a speaker-dependent feature for multi-speaker synthesis.
We design the face encoder based on FaceNet~\cite{schroff2015facenet}, which is composed of several residual blocks consisting of 2D CNNs with pooling layers, dropout, and batch normalization. 
We randomly select one image from the 75 face images and use it as input for the face encoder in order to avoid the extraction of contextual information from a sequence of lip images.

\vspace{2pt}

\noindent\textbf{Decoder.}
Decoder generates mel-spectrograms given the lip and face embeddings.
We use a light version of HiFi-GAN~\cite{kong2020hifi} as the decoder. 
The three main modules of HiFi-GAN form the backbone of our decoder architecture: multi-receptive field fusion (MRF) for a generator, multi-period discriminator (MPD), and multi-scale discriminator (MSD). We use modified configurations of the generator and the discriminator architecture to fit the dimension of the input embedding and the mel-spectrograms, respectively. The upsampling rates of the generator are set to (1, 1, 2), the kernel sizes to (16, 16, 4), and the initial channel dimension to 640. The periods used in the MPD are (2, 3, 5) and the sizes of the convolutional layers in the MSD are (80, 160, 240, 480, 960, 960).

To sum up, the generation process of Facetron is represented as follows:
\begin{align}
    \label{eq:gen}
    {l}_{k} &= \mathbb{E}_{lip}(X_{k}) , ~~~~~~~ f = \mathbb{E}_{face}(X)\\
    {c}_{k} 
    &= concat({l}_{k}, f) \\
    \hat{M} &= \mathbb{G}_{decoder}(C), ~~~~
    C \ni{{c}_{1},..,{c}_{k},...,{c}_{L}} 
\end{align}
The lip embedding $l_{k}$ of the $k$-th frame is extracted from the lip encoder $\mathbb{E}_{lip}(\cdot)$. 
The face embedding $f$ is obtained by the face encoder $\mathbb{E}_{face}(\cdot)$ using a face image $X$.
The input of the decoder $C$ is a sequence of $c_k$'s along the total $L$ frames, each of which is a concatenation of $l_{k}$ and $f$.
Finally, the decoder predicts mel-spectrogram $\hat{M}$.

%% file: 4_training_criterion.tex
\section{Training and inference}
\label{sec:proposed_method}
In this section, we explain a reliable end-to-end training method for the aforementioned modules and a criterion for selecting face embeddings in inference.

\subsection{Training objectives}
The overall loss function of the proposed model is as follows:
\begin{align}
    \label{eq:loss}
    \mathcal{L}_{total} = \mathcal{L}_{Decoder} + 
    \mathcal{L}_{CS} + 
    \mathcal{L}_{Vocoder}, 
\end{align}
where $\mathcal{L}_{Decoder}$, $\mathcal{L}_{CS}$, and $\mathcal{L}_{Vocoder}$ are loss functions for the decoder, the face encoder, and the neural vocoder, respectively.
Note that the lip encoder is trained beforehand because the lip-reading task requires too much time for training.
$\mathcal{L}_{Decoder}$ is used to generate high-quality mel-spectrogram and includes generator loss, discriminator loss, and feature mapping loss as in \cite{kong2020hifi}.

One important issue in training our model is the way of providing speaker-related information to face embeddings.
We introduce prosody information as a target for face embeddings because the speaker identity is strongly related to the speaker's prosody. 
We extract prosody embeddings from a pre-trained prosody encoder in Mellotron~\cite{valle2020mellotron} after fine-tuning it to utilize its general prosody modeling capability.
We adopt a cosine similarity loss $\mathcal{L}_{CS}$ between the prosody embedding and the face embedding as a criterion for the face-prosody matching task:

\begin{gather}
    \label{eq:face}
    \mathcal{L}_{CS} = \mathbf{CS}(f, p), \\ \nonumber
    f = \mathbb{E}_{face}(X), p = \mathbb{E}_{prosody}(M), 
\end{gather}

where $f$ is the face embedding obtained by passing a face image $X$ through the face encoder $\mathbb{E}_{face}(\cdot)$, and $p$ is the prosody embedding obtained by passing mel-spectrogram $M$ through
the fine-tuned prosody encoder $\mathbb{E}_{prosody}(\cdot)$.
For stable training of the face encoder,
we train the face encoder separately from Facetron before the end-to-end training the whole model until the cosine similarity loss does not decrease. 
    
\subsection{Face embedding selection}
\label{face embedding}
During inference, we choose face embeddings using the Inter-to-Intra category distance over I2I method ~\cite{um2020emotional} instead of random selection to accurately represent the target speaker's voice.
First, we calculate the average of each face embedding from the target speaker generated by the face encoder, and also set the average of the face embeddings from a different (negative) speaker whose face embeddings are located closest to the target speaker's embedding. Then, we apply the I2I method as follows: 
\begin{align}
    \mathbf{f}^{I2I}_{T}
    = \frac{1}{2}\arg \max_{\mathbf{f}}
    \frac { \| \mathbf{f}^{Aver}_{N} - \mathbf{f}^{Aver}_{T} \|_2 }
          {\mathbb{E}_{\mathbf{x} \in {X}_T}
           \| \mathbf{f} - \mathbf{x} \|_2 },
\end{align}
where ${f}^{I2I}_{T}$ denotes the selected face embedding from the target speaker. ${X}_T$, ${f}^{Ave}_N$, and ${f}^{Ave}_T$ denote the face embeddings belonging to the target speaker, and average face embeddings of the negative and target speakers, respectively. 
Using this method, Facetron is able to generate high quality mel-spectrograms not only for seen speakers' voices, but also those of unseen speakers'.

%% file: 5_experiment.tex
\section{Experiments and Results}
\label{sec:experiments}
In this section, we compare the performance of the proposed model to that of conventional models~\cite{michelsanti2020vocoder, prajwal2020learning, vougioukas2019videodriven} in terms of synthesized speech quality, linguistic information accuracy, and speaker matching accuracy. 
We use the GRID dataset~\cite{cooke2006audio} for experiments, which contains high-quality audio and facial recordings of 1,000 sentences per speaker, spoken by 33 speakers (17 males, 16 females). 
The sampling rate of the audio signals is 16,000 Hz, and the duration of each sentence is 3 seconds. 
We train and evaluate our model on a small dataset (4 speakers) scenario and a large dataset (29 speakers) scenario to show the effectiveness of our method when trained on different amounts of data. 

\subsection{Small dataset}
\label{exp:small}

\begin{table}[]
\caption{Objective and subjective results of the proposed and reference models for small dataset. The best results are shown in bold.}
\vspace{-5pt}
\label{table:small}
\centering
\resizebox{\columnwidth}{!}{
\begin{tabular}{c|ccc|c}
\Xhline{2\arrayrulewidth}
          & \multicolumn{3}{c|}{Objective} & Subjective \\ \hline\hline
          & NISQA$\uparrow$      & CER$\downarrow$      & WER$\downarrow$      & MOS$\uparrow$        \\ \hline
GAN-based & 2.79             & 9.1                & 18.3               & 1.42           \\
Voc-based & 2.63             & \textbf{6.2}       & \textbf{11}        & 2.19           \\
Lip2wav   & 3.39             & 8                  & 15.4               & 3.57           \\
Facetron  & \textbf{3.47}    & 7.1                & 13.7               & \textbf{3.69}           \\ 
\Xhline{2\arrayrulewidth}
\end{tabular}
}
\end{table}

\begin{table}[]
\caption{Objective and subjective results of the proposed and reference models for large seen/unseen dataset. The (\nth{1}, \nth{2}) and (\nth{3}, \nth{4}, \nth{5}, \nth{6}) rows are results for seen and unseen speaker case, respectively. SMOS of seen speaker case is about speaker similarity and unseen speaker case is for gender similarity.}
\vspace{-5pt}
\label{table:large}
\centering
\resizebox{\columnwidth}{!}{%
\begin{tabular}{c|ccc|cl}
\Xhline{2\arrayrulewidth}
          & \multicolumn{3}{c|}{Objective} & \multicolumn{2}{c}{Subjective} \\ \hline\hline
          & NISQA$\uparrow$     & CER$\downarrow$      & WER$\downarrow$      & MOS$\uparrow$           & SMOS$\uparrow$           \\ \hline
Lip2wav   & \textbf{3.60}    & 7.4               & 11.6               & 3.20                  & 3.53       \\
Facetron  & 3.44             & \textbf{6.9}      & \textbf{10.7}      & \textbf{3.74}         & \textbf{4.16}      \\ 
\hline
Lip2wav   & 3.19             & 12.5               & 24.1                & 2.83                 & 94.56\%    \\      
Facetron  & \textbf{3.24}    & \textbf{7.7}      & \textbf{14.7}       & \textbf{3.87}        & \textbf{99.22\%}           \\ 
\hline
VCVTS   & -             & -               & -                & 2.31                 & -    \\      
Facetron  & -    & -      & -       & \textbf{4.19}        & - \\ 
\Xhline{2\arrayrulewidth}
\end{tabular}bnh
}
\end{table}

\begin{table}[t!]
\caption{Objective and subjective results of the proposed model for disentanglement. }
\vspace{-5pt}
\label{table:disentangle}
\centering
\resizebox{\columnwidth}{!}{%
\begin{tabular}{c|ccc|cl}
\Xhline{2\arrayrulewidth}
          & \multicolumn{3}{c|}{Objective} & \multicolumn{2}{c}{Subjective} \\ \hline\hline
          & NISQA$\uparrow$      & CER$\downarrow$      & WER$\downarrow$      & MOS$\uparrow$           & SMOS$\uparrow$           \\ \hline
Facetron  & 3.41             & 8.3                & 14.3               & 3.57                  & 3.8                \\ 
\Xhline{2\arrayrulewidth}
\end{tabular}
}
\end{table}

A small dataset consists of four speakers (s1, s2, s4, s29), which is the same dataset setup as reference models such as Voc-based~\cite{michelsanti2020vocoder}, GAN-based approach~\cite{vougioukas2019videodriven}, and Lip2Wav model~\cite{prajwal2020learning}. 
We split the small dataset into training, validation, and test sets at ratios of 90\%, 5\%, and 5\%, respectively, which is an identical setup to previous works. 
The output of the Voc-based and GAN-based approach are downsampled to 16 kHz (from 50KHz) to match the frequency range of the others. 
As in our model, we employ a HiFi-GAN vocoder for Lip2Wav.

In lip-to-speech applications, it is crucial to synthesize speech waveforms that convey linguistic information accurately. To evaluate this performance, we performed an automatic speech recognition (ASR) task with the synthesized speech waveforms under the framework used in \cite{chan2016listen}. The ASR model achieves 5.6\% character error rate (CER) and 8.3\% word error rate (WER) for original recorded speech. Table \ref{table:small} shows the recognition accuracy.
Our model achieves the WER of 13.7\%, which is lower than the WER of the GAN-based and Lip2Wav.
Although the Voc-based model shows a higher ASR performance than Facetron, its subjective quality, i.e. mean opinion score (MOS), is significantly lower than Facetron, which means that our model can synthesize natural speech while maintaining the linguistic contents.
Overall, the results show that the speech synthesized by our model contains more accurate linguistic information than the baseline models. 

To objectively measure overall speech quality, we compute non-intrusive objective speech quality assessment (NISQA-TTS)~\cite{mittag2021deep} scores.
Our model outperforms the baseline models with a score of 3.47, showing that it synthesizes much higher speech quality than others when trained on a small dataset.
We also conducted MOS tests on the speech signals synthesized by Facetron and other baseline models. We randomly selected ten audio samples per speaker, that is,  a total of 40 evaluation samples per model. Fifteen subjects participated in the MOS test; they were asked to evaluate the perceptual quality of the audio samples.
Compared to the baseline models, Facetron achieves significantly higher MOS scores than baselines, which is the same result to NISQA.

\subsection{Large dataset}
\label{exp:large}

Large dataset utilizes the full dataset with all 33 speakers.
During training, we excluded four speakers (s1, s2, s4, s29). With these four speakers' data, we evaluated the performance in unseen speakers' conditions.
Table \ref{table:large} shows the results. 
Facetron achieves the MOS scores of 3.74 and 3.87 for seen and unseen data, respectively. Notably, the results are almost identical between the seen and unseen cases, while Lip2Wav demonstrates degraded quality in the unseen case. These results show that Facetron generates high-quality speech signals for both seen and unseen speakers.
In addition, Facetron is better than Lip2Wav in terms of WER; Facetron achieves 10.7\% and 14.7\% WER for seen and unseen speakers while Lip2Wav achieves 11.6\% and 24.1\%, respectively.

We also evaluated our model by conducting subjective speaker and gender recognition on synthesized speech samples. We conducted similarity MOS (SMOS), where subjects were asked to estimate how similar the synthesized and recorded audio samples were from the perspective of speaker and gender, respectively.
The performance of these similarity measurement tasks is directly related to the capability of the face encoder. 
For seen speakers, our model obtains an SMOS score of 4.16 and Lip2Wav obtains 3.53 on the speaker similarity task. 
Therefore, we can say that our face encoder encodes speaker characteristic information for seen speakers much better than Lip2Wav.
For unseen speakers, we performed a gender recognition task because it was not possible to recognize unknown speakers' identities. Our model achieves much higher accuracy than Lip2Wav (99.22\% vs. 94.56\%).

We also tried to compare the performance of our model with those of more recently published models such as VCVTS~\cite{wang2022vcvts}.
However, it was challenging to conduct this comparison because there was no official code available. 
Therefore, we performed a simple MOS test using a small number of speech samples taken from the official VCVTS demo page.\footnote{https://wendison.github.io/VCVTS-demo/}
We downloaded three generated samples of unseen speakers and compared their quality with our generated samples of corresponding speakers.
The speaker split setting of VCVTS is the same as ours as described in Section \ref{exp:large}.
Eleven subjects were guided to evaluate the perceptual quality of the generated audio samples. 
Our model achieved an average MOS score of 4.19 while VCVTS only obtained 2.31, as shown in Table \ref{table:large}. 
Although this was a listening test with only a small number of samples, the result supports the claim that our proposed model has significantly superior performance to VCVTS.

%% file: 6_ablation.tex
\section{Ablation study}
\label{sec:ablation}

\subsection{Disentanglement}
\label{exp:disentangle}
To verify whether linguistic and speaker identity features are successfully disentangled, we randomly synthesized 30 speech samples using lip and identity features from different speakers on the large dataset. 
As shown in Table \ref{table:disentangle}, our model achieves 8.3\% CER, 14.3\% WER, and 3.41 NISQA on objective metrics, and 3.57 MOS and 3.8 SMOS on subjective metrics.
This result means that our model still shows high performance even though we used linguistic and speaker identity features from different speakers.
This verifies that our method is able to effectively extract and disentangle linguistic and acoustic information. 

\vspace{-2pt}
\subsection{Effect of CS loss}
\label{exp:prosody}
\begin{figure}[!t]
\centering
\includegraphics[width=0.9\linewidth]{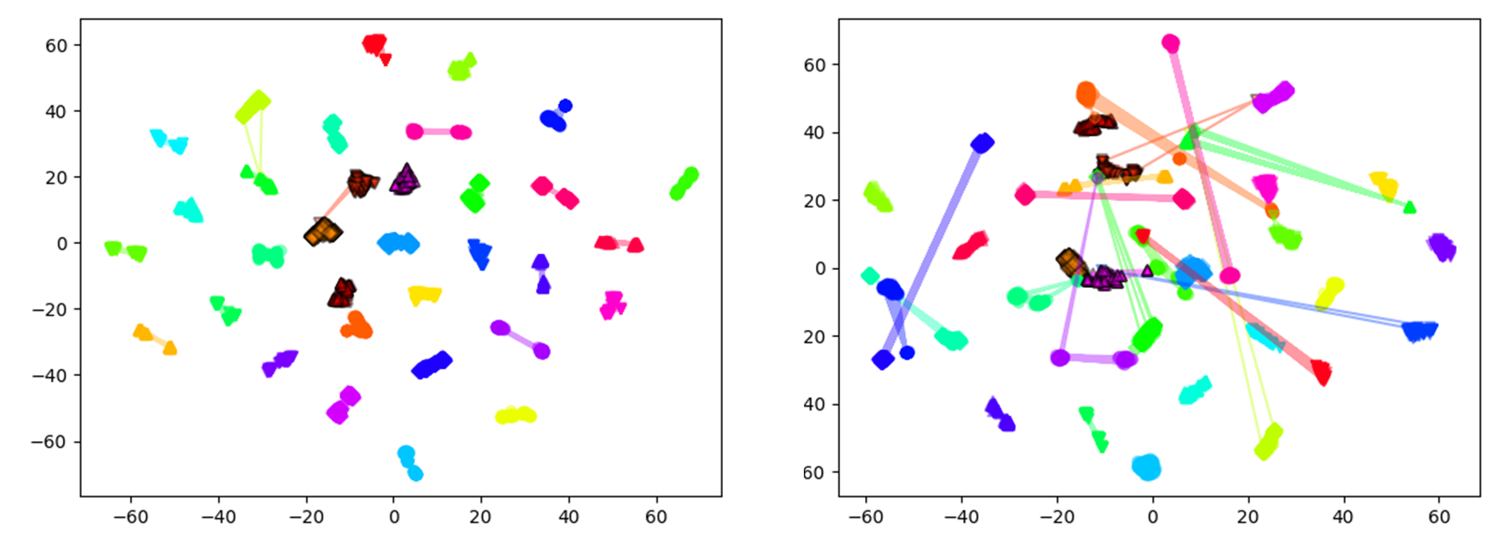}
\caption{\footnotesize Face embeddings extracted from Facetron with CS loss (left) and without it (right). Face embeddings of the same speaker are represented using the same type of color and marker and connected by a translucent line. Unseen speaker's face embedding has a black border.}
\label{fig:style_embedding}
\vspace{-5pt}
\end{figure}

The face encoder learns encoding speaker-related information from face images by training with the cosine similarity (CS) to prosody embeddings, as described in Section \ref{sec:proposed_method}. Figure \ref{fig:style_embedding} illustrates the t-SNE plots to demonstrate the advantage of using CS loss.
Face embeddings generated by Facetron with CS loss form clear speaker clusters while embeddings from the model without CS loss form distributed clusters. 
The effectiveness of CS loss is verified when synthesizing unseen speakers' speech. 
Without CS loss, the model can fail to synthesize a consistent speaker identity in a single speech sample.
For example, the front part of the sentence can be spoken in a male voice while the back part can be spoken in a female voice, which is not desireable.
Examples of these samples has been uploaded to the demo page. 

%% file: 7_conclusion.tex
\section{Conclusion}
\label{sec:conclusion}

In this paper, we proposed Facetron, a model for synthesizing speech from images of facial movements. Our model combines a lip embedding that contains linguistic information and a face-driven embedding that contains speaker information and is trained using a cross-modal learning framework.
Experiments show that our model is able to synthesize more natural and intelligible speech compared to previously proposed methods.

\section{Acknowledgements}
\label{sec:acknowledge}

This research was supported by the Yonsei Signature Research Cluster Program of 2022 (2022-22-0002).